\pdfoutput=1

\documentclass[11pt]{article}

\usepackage[preprint]{acl}

\usepackage{times}
\usepackage{latexsym}
\usepackage{graphicx}
\usepackage{multirow}
\usepackage{booktabs}
\usepackage{amsmath,amssymb}
\usepackage{soul}
\usepackage[T1]{fontenc}

\usepackage[utf8]{inputenc}

\usepackage{microtype}

\usepackage{inconsolata}

\title{An Automatic Prompt Generation System for Tabular Data Tasks}

\author{Ashlesha Akella\\
  IBM Research, India\\
  \texttt{ashlesha.akella@ibm.com}
  \\\And
  Abhijit Manatkar\\
  IBM Research, India\\
  \texttt{abhijitmanatkar@ibm.com}
  \\\And
  Brij Chavda\\
  IBM Research, India\\
  \texttt{brijkumar.chavda@ibm.com}
  \\\AND
  Hima Patel\\
  IBM Research, India\\
  \texttt{himapatel@in.ibm.com}
  }

\begin{document}
\maketitle
\begin{abstract}

Efficient processing of tabular data is important in various industries, especially when working with datasets containing a large number of columns. Large language models (LLMs) have demonstrated their ability on several tasks through carefully crafted prompts. However, creating effective prompts for tabular datasets is challenging due to the structured nature of the data and the need to manage numerous columns. This paper presents an innovative auto-prompt generation system suitable for multiple LLMs, with minimal training. It proposes two novel methods; 1) A Reinforcement Learning-based algorithm for identifying and sequencing task-relevant columns 2) Cell-level similarity-based approach for enhancing few-shot example selection. Our approach has been extensively tested across 66 datasets, demonstrating improved performance in three downstream tasks: data imputation, error detection, and entity matching using two distinct LLMs; Google flan-t5-xxl and Mixtral 8x7B.

\end{abstract}

\section{Introduction}

Recent advancements in pre-training large language models have paved the way for prompt-based and in-context learning \cite{brown2020language,raffel2020exploring}, providing an efficient approach to tackle a wide range of tasks. Generating a suitable prompt is particularly important when harnessing pre-trained LLMs for tabular downstream tasks. Unlike natural language sentences, tabular data necessitates specific formatting, column preferences, and in-context examples. In the domain of tabular data downstream tasks, prompts are frequently customized for each dataset and specific downstream tasks to ensure consistent and efficient performance.  A few studies \cite{narayan2022can,zhang2023large} have demonstrated this approach, wherein prompts are crafted for a given dataset and task by manually selecting columns and relevant in-context (few-shot) examples. Additionally, recent research has introduced automated methods that target specific components of prompt such as in-context examples \cite{huh2023pool}. This leads us to two pivotal questions; Firstly, what are the components or parts of the prompt that  significantly impact performance in tabular tasks? Secondly, how can we devise automated methods for the components to generate prompts that can be efficient and effective for tabular data tasks?

\begin{figure}
    \centering
    \includegraphics[width=0.8\linewidth]{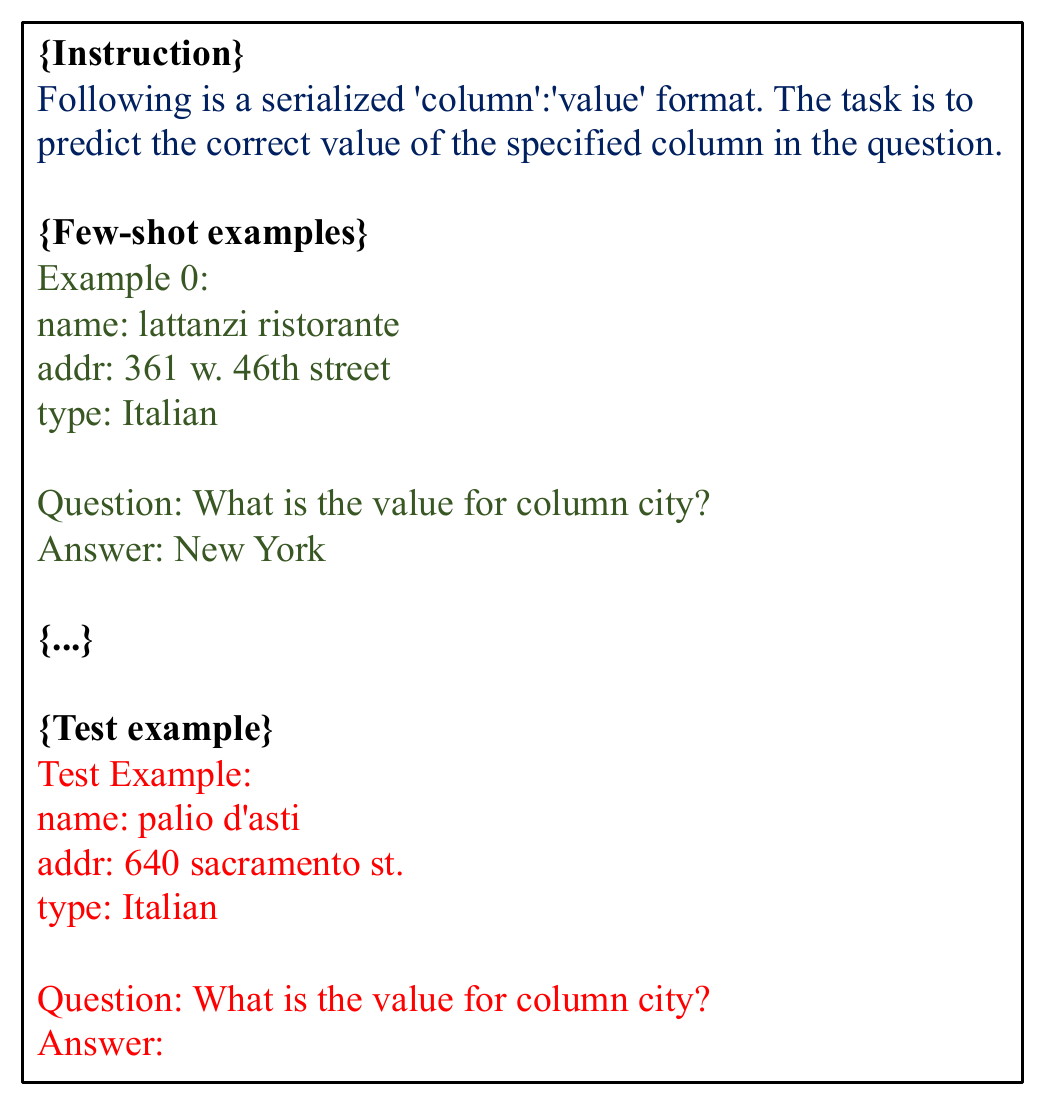}
    \caption{Example Prompt Template for Data Imputation task}
    \label{fig:prompt-template}
\end{figure}

In tabular data tasks, a vanilla prompt typically includes a row of information where each column name is paired with its respective value, for example: \emph{Brand: Dell; Price: \$349.00; Feature: Newest Dell Inspiron}. However, providing all column information of a row may introduce unnecessary details, redundancy, and noise, while also inefficiently using input tokens, potentially leaving inadequate space for additional crucial information such as few-shot examples. Our empirical analysis reveals that carefully selecting columns to be included in the prompt improves task performance, aligning with prior research findings by \cite{narayan2022can,zhang2023large}. In addition to choosing the right columns, we noticed a significant improvement in performance when we carefully arranged these column details in the prompt. This underscores the importance of both selecting the columns and arranging them in a specific order. Particularly, dealing with large datasets with many columns poses a practical challenge.

Furthermore, our study demonstrates a performance difference between traditional few-shot example selection methods developed for natural language (NL) focused tasks and our proposed cell-level similarity few-shot (CLFS) example selection approach for tabular data. The NL-based method serializes a row into a sentence and selects few-shot examples based on similarity, potentially losing relevant information by imposing a sentence structure on the tabular data during serialization. In contrast, the proposed CLFS method considers each cell's information independently with the aim of selecting few-shot examples at a cell level similarity.
 
To this end, we introduced an auto-prompt generation system designed to be compatible with various LLMs without the need for extensive training. This system introduces a novel approach emphasizing two essential elements:
\begin{enumerate}
    \item The identification and sequencing of task-relevant columns facilitated by a Reinforcement Learning-based algorithm.
    \item A few-shot selection approach based on cell-level similarity.
\end{enumerate}

\section{Related Work}

Recent studies on tabular data tasks have demonstrated that LLMs can effectively tackle data wrangling tasks, through different strategies, including pre-training and fine-tuning \cite{gong2020tablegpt,iida2021tabbie,somepalli2021saint,wang2021tuta,tang2020rpt}, prefix-tuning \cite{vos2022towards} and prompt learning \cite{liu2022structured,chen2023trompt,zhang2023jellyfish}. However, these training methodologies pose significant computational demands and exhibit high time complexity. Furthermore, many of these approaches require adjustments to the model parameters, a process that proves impractical for black-box language models like ChatGPT.

Previous research has achieved success in developing methods for generating prompts for tabular data tasks \cite{narayan2022can,zhang2023large}. However, a challenge exists due to the reliance on manual processes to select columns and few-shot examples. This manual approach becomes especially difficult when dealing with large datasets with many columns. The study by \cite{huh2023pool} propose a few-shot selection method that utilizes an embedding of a row transformed into a natural language sentence, raising questions about integrating tabular structure-aware few-shot selection methods for tabular data.

This paper discusses the importance of selecting and organizing columns with empirical results, followed by an overview of two auto-prompt generation systems. Extensive testing was carried out across 3 different tabular data tasks; Data Imputation (DI), Error Detection (ED), and Entity Matching (EM), utilizing 66 datasets using two models: \textbf{Google flan-t5-xxl} (11B parameters) \cite{chung2022scaling} and \textbf{Mixtral 8x7B} (47B parameters) \cite{jiang2024mixtral} \footnote{Specifically, we use the GPTQ \cite{frantar2022gptq} quantized version of the model available at \url{https://huggingface.co/TheBloke/Mixtral-8x7B-Instruct-v0.1-GPTQ}. Mixtral 8x7B is a Sparse Mixture of Experts (SMoE) Model consisting of 8 feedforward blocks (i.e. experts) at each layer. For each token, at each layer, 2 out of 8 experts are selected for inference which results in a total of 13B active parameters.}.

\section{Motivation}

Tabular data, especially with wider datasets, presents a challenge when inputting all column details into LLMs for row-level tasks (e.g., Data Imputation, Error Detection). It becomes evident that selecting columns is crucial for optimizing the performance of LLMs. 

\begin{figure}[h]
    \centering
    \includegraphics[width=\linewidth]{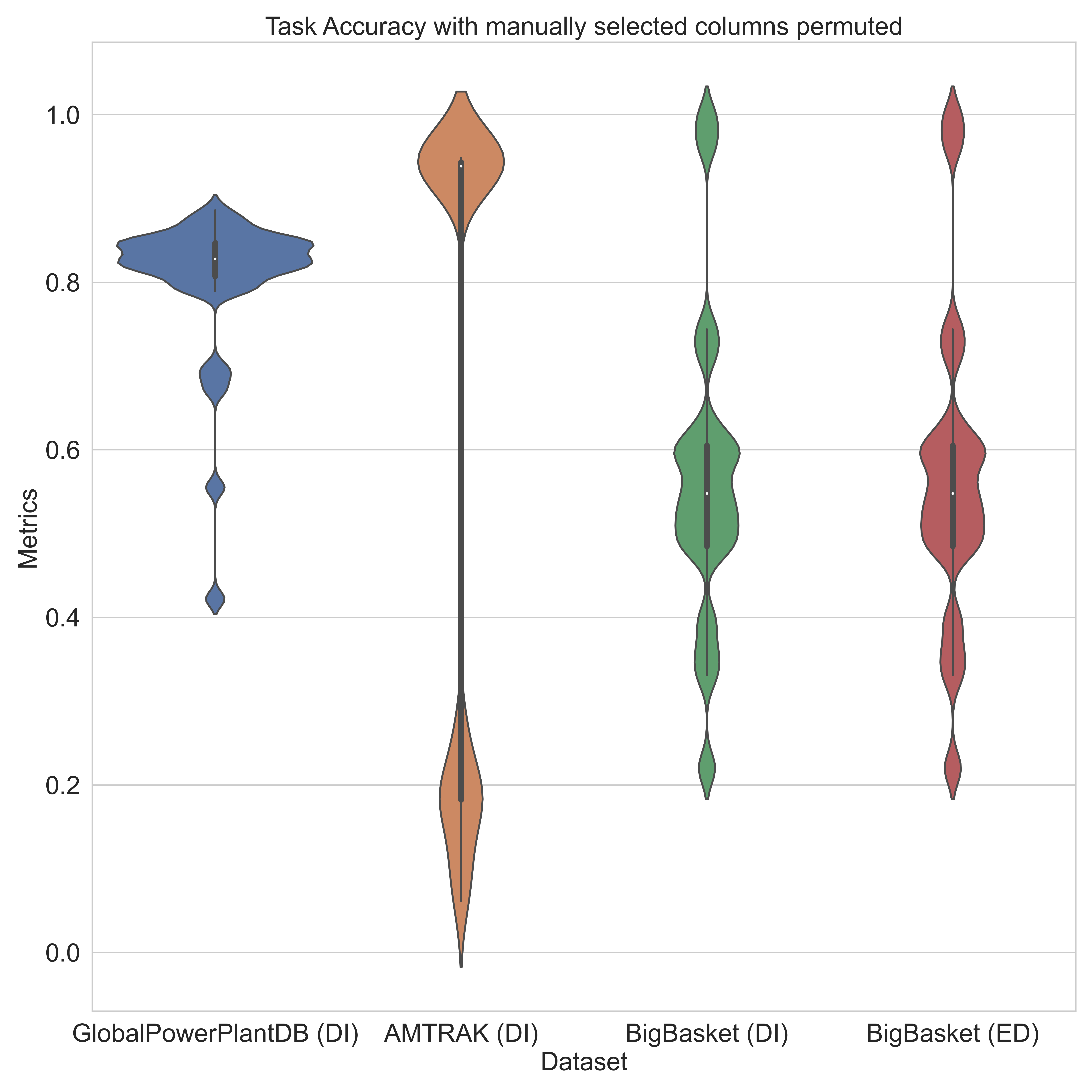}
    \caption{Variations in accuracy across different combinations and permutations for manually selected columns for Data Imputation (DI) and Error detection (ED). We collected accuracies for all possible permutations of the selected columns (per dataset and per task) and visualized the distributions of accuracies.}
    \label{fig:col_permuted}
\end{figure}

In addition to choosing columns, we conducted a study to explore the impact of column arrangement on downstream tasks performance. In this experiment, we manually selected specific columns for a dataset and downstream task and then created prompts using a template (see Figure \ref{fig:prompt-template}). Figure \ref{fig:col_permuted} demonstrates the significant effect of different column orders on accuracy. For instance, when examining the Data Imputation task on the AMTRAK dataset, a wide range of accuracies was observed, varying from 0.06 to 0.95 for the manually selected columns but in different orders and combinations. This shows that using an optimal sequence of subset columns is crucial while generating a prompt for tabular data tasks. 

This can be considered as solving a sequential decision-making problem. This study uses Reinforcement Learning (RL) to optimize column selection and ordering in order to improve accuracy and performance in tabular data tasks.

Another component, namely few-shot example selection, becomes critical in light of the improved performance shown by LLMs when provided with a small set of illustrative examples (few-shot examples) in the prompt. Selecting these examples is extensively studied in the field of Natural Language tasks  \cite{ma2023fairness,liu2021makes} and these methods typically retrieve similar examples from a pool using similarity metrics such as BM25 \cite{robertson1995okapi}  or cosine-similarity calculated over task specification embeddings obtained from encoder-only models like BERT \cite{devlin2018bert} and RoBERTa \cite{liu2019roberta}, etc. Our exploration reveals that a modified approach for calculating the similarity measure leads to improved performance. To address this, our work proposes a new \textbf{Cell-Level Similarity Measure} for retrieving in-context examples that takes into account the semantic similarity of individual cells in a row, and is seen to outperform natural language inspired baselines. 

We evaluated the performance of the proposed system on 3 downstream tasks Data Imputation (DI), Error Detection (ED) and Entity Matching (EM). More details on the downstream as provided in Appendix \ref{downstream_tasks}. 

\section{Method and Implementation}

\begin{figure*}[ht]
    \centering
    \includegraphics[width=1\textwidth]{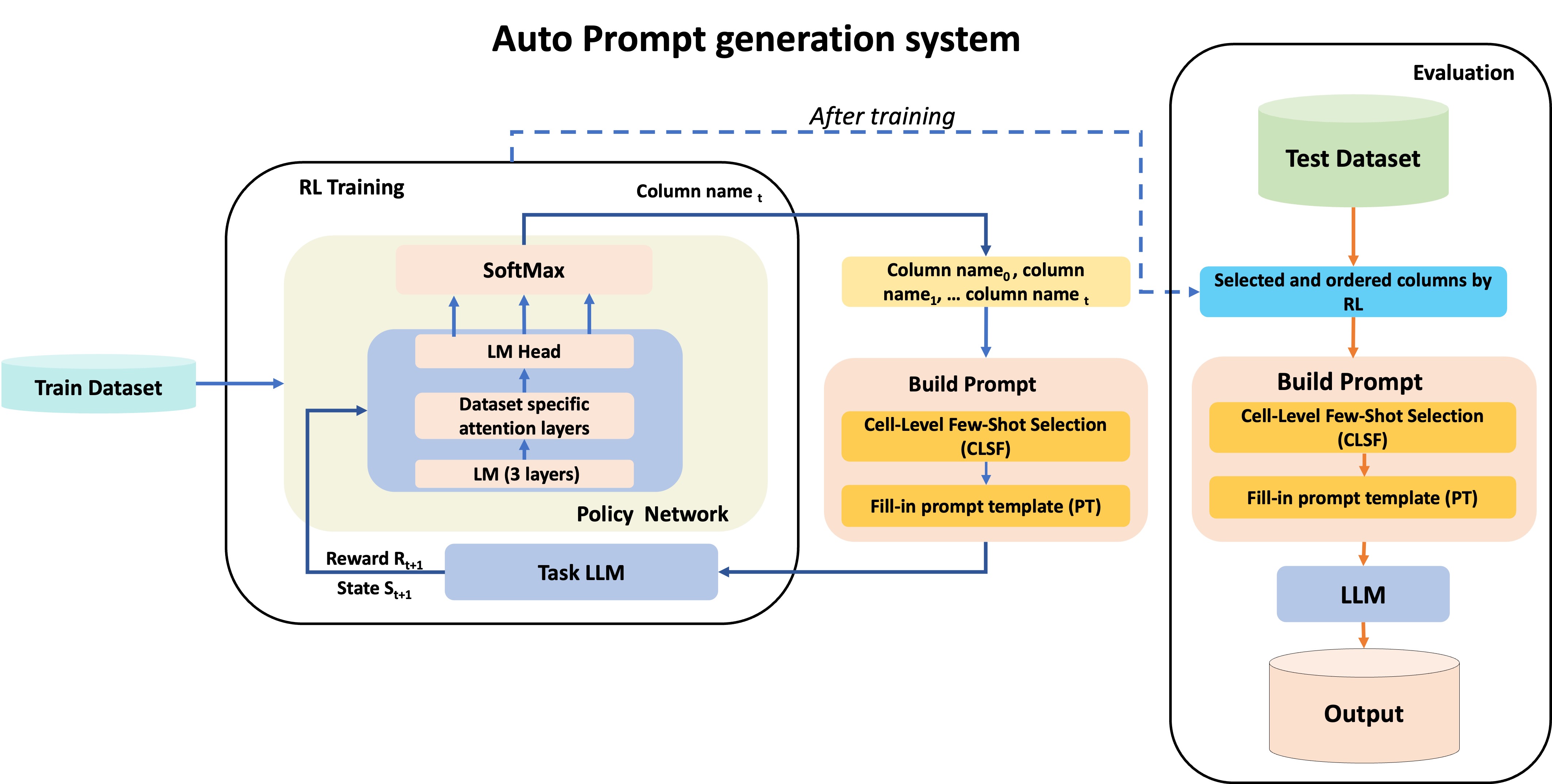}
    \caption{The architecture comprises three modules: RL agent Training Module for Column Selection, Build Prompt Module and Evaluation.}
    \label{fig:arch}
\end{figure*}

In this section, we explain the methodology of developing our auto-prompt generation system. We begin by describing the overall architecture of the system, followed by description of individual components.

\subsection{Architecture}
The architecture of our system, as seen in Figure \ref{fig:arch}, comprises of three modules: RL agent training for Column Selection (RLCS), Build Prompt module, and Evaluation. The RL agent is trained to generate a sequence of column names. The Build Prompt module contains Cell-level Similarity based Few-shot selection (CLFS) to select few-shot examples and the Prompt Template module which fills in a predefined prompt template with selected column information from the test sample and the selected few-shot examples. Once an RL agent is trained, we obtain an optimal sequence of selected columns from the final model which is used during evaluation.

\subsection{Reinforcement Learning based Column Selection: RLCS}
\label{columns-selection}

The process of choosing columns as a sequence can be seen as a decision-making procedure that can be represented as a Markov Decision Process (MDP). This includes a set of states $S$, a set of actions $A$, a transition function $P:S \times A \rightarrow S $ and a reward function $R : S \rightarrow {\mathbb{R}}$. The objective for the RL agent is to maximize its expected cumulative reward defined by $\mathcal{R}=\mathop{\mathbb{E}}[\sum_{t=0}^T\gamma^t r_t]$, where $r_t$ denotes reward at step $t$ and $\gamma \in [0, 1]$ signifies a discounting factor for rewards over time steps. We use the \textbf{Soft Q-Learning} algorithm \cite{haarnoja2017reinforcement,guo2021efficient} to train the network with the goal of maximizing cumulative rewards within an episode. 

\subsubsection{State and Action Representation}

An initial state is defined as $s_0=d_0$, where $d_0$ is a brief description of the dataset (e.g: $d_0$ = `This dataset contains products on the Bigbasket website'). At each time step $t$, the RL agent selects one column name $a_t$ from the action space $A$. The transition function $P(s_t, a_t) = s_{t+1}$ appends the selected column to the previous state, i.e., $s_{t+1}$ = $\oplus(s_t,a_t)$. There exists termination criteria based on the number of columns chosen.

\subsubsection{Policy Network}

We use an attention-based architecture in the policy network for the RL agent. The first three layers of the policy network are taken from a small LLM and kept frozen. Specifically, we utilize the initial 3 layers of GPT-2 \cite{radford2019language}. This is followed by a \textit{trainable} multi-headed attention layer with two heads, followed by the LM head layer from GPT-2. The initial three layers and the LM head layer are taken from the smallest version of GPT-2 with 124M parameters. Let $\mathcal{M}$ represent the policy network and $\mathcal{M}\left(y \mid s, {\theta}\right)$ be the logit value of policy network for a token $y$ given state $s$ and weight parameters $\theta$. During training, only the multi headed attention layer is trained while all other layers remain frozen. For every column, $col \ name_{i}$ we calculate the mean of logits for all $n_i$ tokens $\{ y^i_0, y^i_1, \dots y^i_{n_i-1}\}$ in the name of that column, yielding  $q_{s, \textit{col name}_{i}}$. This computation is performed for all columns ($\forall i \in \{0,1,\dots N-1\}$), resulting in the vector of Q-values $\mathbf{q_s}$. Subsequently, a softmax function is applied to $\mathbf{q_s}$ to obtain a probability distribution over the columns.  During exploitation, the action $a_t$ is $\arg \max$ over this distribution. During exploration (i.e. while training), $a_t$ is obtained by sampling from this distribution.

\begin{equation}
    A = \{\textit{col name}_{0},\textit{col name}_{1}, \dots \textit{col name}_{N-1}\} 
\end{equation}
\begin{equation}
    \mathbf{q_s} = \{ q_{s, \textit{col name}_{0}} \dots q_{s, \textit{col name}_{N-1}} \}
\end{equation}
\begin{equation}
    q_{s, \textit{col name}_{i}} =  \frac{1}{n_i} \sum_{k=0}^{n_i - 1} \mathcal{M}\left(y_k^{i} \mid s, \theta \right)   
\end{equation}
\begin{equation}
    a_{t} = \underset{a \in A}{\arg \max } \left(softmax (\mathbf{q_s})\right ) 
\end{equation}
Here $N$ is the number of columns, $n_i$ represents the total number of tokens of column $\textit{col name}_{i}$.

At each timestep, a new column name is added to the list of selected columns. Based on these selections, CLFS chooses few-shot examples that are then sent to the PT module (shown in Figure \ref{fig:arch}) as input for Task-LM. The agent receives a reward at each timestep $t$ following:
\begin{equation}
    r_t = 
    \begin{cases}
        20 - 3t & \text{if Task-LM matches the } \\
                & \text{expected output} \\
        -0.5 & \text{otherwise}
    \end{cases}
\end{equation}

Figure \ref{fig:reward_plot} in the Appendix illustrates that the RL-based approach can identify an optimal sequence of column sets across training episodes. The columns chosen by RL and those selected manually for each dataset and task can be found in Appendix \ref{selected_cols}.

\begin{table*}[ht]
    \centering
    \resizebox{\linewidth}{!}{
    \begin{tabular}{ccccccccc}
        \toprule
        \multirow{3}{*}{\textbf{Task}} & \textbf{Dataset} & \textbf{Baseline} & \textbf{MCS-RFS} & \textbf{MCS-NFS} & \textbf{MCS-CLFS} & \textbf{RLCS-CLFS} &\textbf{Baseline} & \textbf{RLCS-CLFS} \\
        & \#columns & flan-t5-xxl & flan-t5-xxl& flan-t5-xxl & flan-t5-xxl & flan-t5-xxl & Mixtral 8x7B & Mixtral 8x7B\\
        & & & &  &  & (Ours) & & (Ours)\\
        \midrule 
        \multirow{4}{*}{DI} & Restaurant (6) & $0.75 \pm 0.02$ & $0.76 \pm 0.02$ & 0.75 & 0.77 & \textbf{0.82} & 0.92 & \textbf{0.97}\\
                            & BigBasket (14) & $0.32 \pm 0.12$ & $0.75 \pm 0.00$ & 0.86 & \textbf{0.93} & 0.92 & 0.91 & \textbf{0.92}\\
                            & GlobalPowerPlant (40) & $0.61 \pm 0.06$ & $0.75 \pm 0.00$ & 0.85 & 0.85 & \textbf{0.90} & 0.84 & 0.84\\ 
                            & AMTRAK (86) & $0.57 \pm 0.11$ & $0.91 \pm 0.00$ & 0.93 & 0.92 & \textbf{0.98} & 0.61 & \textbf{0.81}\\ 
        \midrule   
        \multirow{4}{*}{ED} & Adult (15) & $0.50 \pm 0.00$ & $0.62 \pm 0.02$ & \textbf{0.96} & 0.89 & 0.95 & 0.54 & \textbf{0.78}\\
                            & Hospital (23) & $0.49 \pm 0.00$ & $0.70 \pm 0.00$ & 0.56 & 0.57 & \textbf{0.85} & 0.36 & \textbf{0.81} \\ 
                            & Global PowerPlant (40) & $0.33 \pm 0.11$ & $0.34 \pm 0.00$ & 0.42 & 0.52 & \textbf{0.83} & 0.90 & \textbf{0.95} \\ 
                            & BigBasket (14) & $0.39 \pm 0.12$ & $0.38 \pm 0.00$ & 0.39 & 0.39 & \textbf{0.87} & 0.37 & \textbf{0.95} \\
        \midrule                 
        \multirow{4}{*}{EM} & Fodors-Zagats (14)& $0.86 \pm 0.01$ & $0.97 \pm 0.02$ & 0.96 & 0.93 & \textbf{1.00} & 0.92 & \textbf{0.99}\\
                            & DBLP-GoogleScholar (10)& $0.69 \pm 0.02$ & $0.84 \pm 0.04$ & 0.74 & 0.81 &  \textbf{0.85} & 0.83 & \textbf{0.85}\\ 
                            & beers (10)& $0.71 \pm 0.03$ & $0.84 \pm 0.08$ & 0.81 & 0.81 &  \textbf{0.89} & 0.88 & \textbf{0.91}\\
                            & Walmart-Amazon (12)& $0.54 \pm 0.02$ & $0.79 \pm 0.04$ & 0.87 & 0.87 & \textbf{0.87} & 0.91 & 0.91\\
        \bottomrule
    \end{tabular}}
    
\caption{Performance of Data Imputation (DI), Error Detection (ED) and Entity Matching (EM) tasks under 5 varied conditions on \textbf{Google flan-t5-xxl (11B)} model and 2 conditions with \textbf{Mixtral 8x7B}. \#columns shows the number of columns in the Dataset. As metrics, accuracy is used for DI and F1-macro used for ED and EM. Baseline and MCS-RFS experiments are for 3 different seeds, where accuracy is $\emph{avg}\pm\emph{std}$ }
\label{tab:result_table_t5_xxl}
\end{table*}

\subsection{Cell-Level Similarity Measure based Few-shot Selection: CLFS}
\label{CLFS}
The proposed Cell-Level Similarity Measure uses an embedding technique to preserve the semantic content of each cell in a row. A pool of examples, denoted as $P$, is available for selecting a few-shot examples. Two methods were explored to understand the performance difference between Natural Language-based (NL) approach and the proposed CLFS approach.

In the NL-based approach, an embedding for each row is generated by serializing it using a template (as discussed in section \ref{sec:prompt-template}) and then encoding it into a latent space using model $B$ (typically an encoder-only transformer model). For every test sample, cosine similarity ($sim_{NL}$) was calculated between the test sample $r_t$ and all samples from the pool $P$, and then top k samples from $P$ with the highest similarity scores with the test sample are chosen as few-shot examples for that test sample.
\begin{equation} \label{eq:simnl}
    sim_{NL}(r_{t}, r_{x}) = B(ser(r_{t}))^T B(ser(r_{x}))
\end{equation}
where $ser(r)$ serializes row $r$, $r_t$ and $r_x$ are the test sample and a sample from $P$ respectively.

This method requires presenting a row of tabular data as a serialized string of text which is then embedded using a Language Model trained on natural language data. The presentation and the encoding model in this method treat the table row like a string of natural language text, which can result in sub-optimal embeddings because of information loss from a representational mismatch. The proposed CLFS method embeds each cell of the row independently of other cells, and then computes the similarity $sim_{CL}$ between a test sample $r_t$ and a sample $r_x$ from pool $P$ as the average of similarities between corresponding cells.
\begin{equation} \label{eq:simcl}
sim_{CL}(r_t, r_x)=\frac{\sum_{c \in C} B(r_t[c])^T B(r_x[c])}{|C|}
\end{equation}
where $C$ is the set of columns and $r[c]$ gives the value for the cell at the intersection of column $c \in C$ and row $r$.

\subsection{Prompt template}
\label{sec:prompt-template}
The prompt template for tabular data wrangling tasks includes a brief description of the serialization, followed by serialized few-shot examples and a test example. An illustrative example of the prompt template is presented in Figure \ref{fig:prompt-template}. The serialization of both the few-shot examples and the test example follows 
\begin{equation*}
{F}_i^r = \Downarrow_{n=1}^N Example\ {n} : \oplus \Downarrow_{j=1}^c h_j^n \oplus: \oplus v_{n, j}^n \oplus{} 
\end{equation*}
\begin{equation*}
    {S}_i^r = {F}_i^r \Downarrow Test Example : \oplus \Downarrow_{j=1}^c h_j^i \oplus: \oplus v_{j}^i \oplus{} 
\end{equation*}

For $i^{th}$ test row, ${F}^r_i$ is serialized $n^{th}$ few-shot, $N$ is the total number of few-shot examples, $c$ is the number of columns, $h_j^n$ is the $j^{th}$ column name and $v_{j}^n $ is the value of column $j$ of $n^{th}$ few-shot. $\Downarrow$ is the new line operator and $\oplus$ is the concatenation operator.


\section{Datasets}
We gathered datasets from various sources like Kaggle \footnote{https://www.kaggle.com}  and Open ML \footnote{https://openml.org}, ensuring datasets containing numerous columns (upto 120 columns) across different domains. All the datasets gathered are in the format Comma Seperated Values (CSV) files. For the data imputation task, specific columns were chosen for imputing values across all rows within those columns. Real-world databases commonly have both syntactic and semantic errors \cite{chu2013holistic, heidari2019holodetect, mayfield2010eracer}. In the error detection task, selected columns in the datasets were introduced with errors: around 25\% of cell values were replaced with out-of-domain strings for semantic errors (e.g., `Stationer' in the `County name' column), while approximately 25\% of cell values within a specific column had random letter additions introduced as syntactic errors. For entity matching tasks, we used datasets previously studied in literature \cite{mudgal2018deep}.

\section{Experimental Results}

This section describes the experiments carried out in our study. We compared 5 different conditions including our proposed system to highlight the importance of each of the components. These experiments included 12 datasets across 3 tasks using Google flan-t5-xxl as shown in Table \ref{tab:result_table_t5_xxl}. On observing similar trends with Mixtral 8x7B as with flan-t5-xxl, we only report results for our system and the baseline with Mixtral 8x7B.

\begin{enumerate}
    \item \emph{Baseline}: For the baseline method, no column selection is done and data from all columns in included in the prompt. Columns are permuted in the order in which they appear in the dataset and few-shot examples are chosen randomly. We conduct experiments with three different seeds.
    \item \emph{Manual Column Selection and Random Few-shot examples (MCS-RFS)}: To assess the efficacy of carefully chosen manual columns, we conducted experiments by manually selecting columns and selecting random few-shot examples using three different seeds, while keeping the column order consistent across all seeds.
    \item \emph{Manual Column Selection and NLP Few-shot Selection (MCS-NFS)}: This method seeks to assess the performance of the cosine similarity few-shot selection method used in Natural Language tasks. For selecting few-shot examples, the $sim_{NL}$ similarity metric (\ref{eq:simnl}) is used.
    \item \emph{Manual Column Selection and Cell-Level Similarity Few-shot selection (MCS-CLFS)}: This method seeks to assess the impact of using the cell level similarity metric $sim_{CL}$ (\ref{eq:simcl}) for selecting few-shot examples. In this method and the previous one, columns are selected manually while keeping the column order consistent across methods.
    
\end{enumerate}
Additionally, the comparison between baseline and auto-prompt generation results was conducted across 66 datasets (see Appendix Tables \ref{tab:DI_all_datasets}, \ref{tab:ED_all_datasets} and \ref{tab:EM_all_datasets}). Each of the studied datasets are partitioned into train, validation and test splits. The train split is used for training the RL-based column selection agent. The validation split is used as the pool $P$ for selecting few-shot examples and metrics are reported on the test split. For the settings which use few-shot example selection, \texttt{all-distilroberta-v1} from the SentenceTransformers library \cite{reimers-2019-sentence-bert} is used as the encoding model $B$.

\section{Conclusion and Future Work}

The results indicate that our proposed system significantly outperforms the baseline as well as methods based on combinations of manual column selection, random few-shot selection and natural language based few-shot selection. Our research underscores the efficacy of an auto-prompt generation system in enhancing tabular data tasks across various datasets and tasks using Large Language Models. Manual column selection and sequencing, found to be a cumbersome process, necessitates an automatic method, a gap which is suitable filled by our RL-based algorithm. Our proposed cell-level similarity measure exhibits improved performance compared to the NL-based few-shot selection method. Overall, the auto-prompt generation system showcases versatility and generalizability across diverse tasks and datasets, providing a streamlined solution for efficient tabular data tasks.

As part of future work, the following directions of study may be worthwhile:
\begin{enumerate}
    \item Expanding the automatic prompt-generation system to more row-level downstream tasks and analyzing its performance.
    \item Training of a unified model for column selection and sequencing across tabular datasets.
    \item Identifying other parts/components of the prompt that can be automatically optimized for tabular tasks.
\end{enumerate}

\bibliography{custom}

\appendix

\section{Appendix}
\label{sec:appendix}

\subsection{Reinforcement Learning Parameters}
\label{sec:rl-params}

The hyperparameter settings for reinforcement learning employed in this study are delineated in Table \ref{tab:rl_params}.

\subsection{Reinforcement Learning Reward Graph}

\begin{figure*}[ht]
    \centering
    \includegraphics[width=1\textwidth]{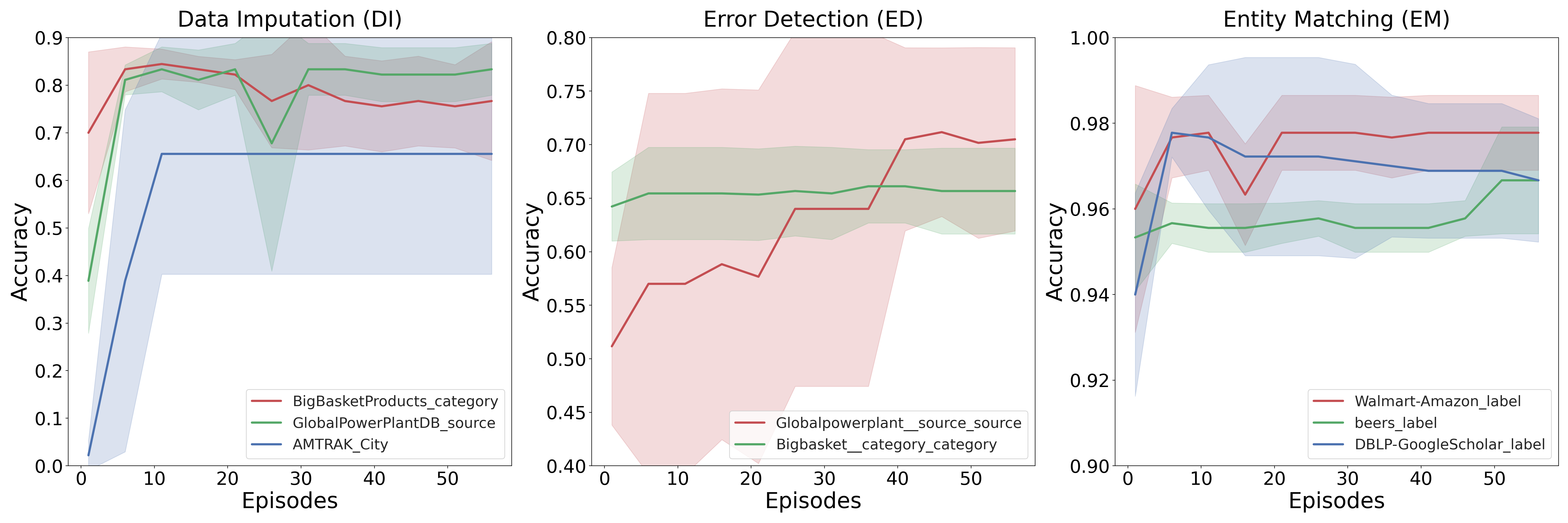}
    \caption{The plot shows, reward accumulated by the RL-agent while undergoing training for each episode. The solid lines represent the average, and the shaded areas depict the highest and lowest test accuracy across 3 different seeds.}
    \label{fig:reward_plot}
\end{figure*}

Figure \ref{fig:reward_plot} illustrates that the RL-based approach can identify an optimal sequence of column sets across training episodes.

\subsection{Manually selected columns and RL selected columns}
\label{selected_cols}
Table \ref{tab:sel_cols} shows the columns that were manually selected and those selected and sequenced by the RL algorithm for 12 datasets across three tasks.

\subsection{Downstream Tasks}
\label{downstream_tasks}
In the assessment of an Auto-prompt generation system for tabular datasets, we performed three distinct downstream tasks: Data Imputation, Error Detection, and Entity Matching. 

\begin{itemize}
    \item Data Imputation (DI): DI entails predicting missing values in a given column and row. For instance, if a dataset for restaurants has some missing values in the "state" column, the data imputation task involves predicting the value of "state" based on other details for that specific row. 

    \item Error Detection (ED): ED focuses on identifying errors within a given row. As an example, consider a dataset where an error is present in the "City" column with the value "Computer." 
    
    \item Entity Matching(EM): involves comparing two rows to determine if they match semantically. For instance, when comparing two CSV files containing details of products from different ecommerce websites, this task aims to establish whether there is a match between each product listed in one CSV file with those listed in another.

\end{itemize}

\subsection{Overall Results}

Table \ref{tab:DI_all_datasets}, Table \ref{tab:ED_all_datasets} and Table \ref{tab:EM_all_datasets} display the extensive results of the proposed auto-prompt generation system across three tasks: data imputation, error detection, and entity matching. The system utilized a diverse set of 66 datasets.

\begin{table}[ht]
    \small
    \begin{tabular}{|c| c |c|}
        \toprule
        Parameter Name & Parameter Value  \\
        \hline
        Discount factor $\gamma$ & 0.6 \\
        \hline
        Learning rate & $1e-4$ \\
        \hline
        Batch size & 200 \\
        \hline
        Number of episodes & 60 \\
        \hline
        Exploration fact $\epsilon$ & 0.4  \\
        \hline
        Max replay buffer size & 3000 \\
    \bottomrule
    \end{tabular}
    \caption{Reinforcement Learning Hyper Parameters}
    \label{tab:rl_params}
\end{table}

\begin{table*}[ht]
    \centering
    \small
    \begin{tabular}{cccccc}
    \toprule
    \textbf{Dataset} & \textbf{Target Column} & \textbf{Baseline} & \textbf{Ours} & \textbf{Baseline} & \textbf{Ours} \\
    (\#columns) & & flan-t5-xxl (11B) & flan-t5-xxl (11B) & Mixtral (8x7B) & Mixtral (8x7B) \\
    \midrule
    Airline Dataset (15) & Country Name  & 0.78 & \textbf{0.97} & 0.92 & \textbf{0.99} \\
    Airline Dataset (15) & Airport Continent & 0.87 & \textbf{1.00} & 0.67 & \textbf{1.00} \\
    Airline Dataset (15) & Airport Country Code  & 0.68 & \textbf{0.90} & 0.99 & \textbf{1.00} \\
    Airline Dataset (15) & Continents  & 0.91 & \textbf{1.00} & 1.00 & 1.00 \\
    customer support tickets (17) & Ticket Type  & 0.21 & \textbf{0.21} & 0.08 & \textbf{0.20} \\
    customer support tickets (17) & Ticket Priority  & 0.16 & \textbf{0.27} & 0.14 & \textbf{0.25} \\
    customer support tickets (17) & Ticket Subject & 0.01 & \textbf{0.06} & 0.01 & \textbf{0.05} \\
    finance sentiment analysis (2) & Sentiment  & 0.51 & 0.51 & 0.68 & \textbf{0.70} \\
    flipkart ecommerce (15) & product\_category\_tree  & 0.12 & \textbf{0.48} & 0.01 & \textbf{0.31} \\
    flipkart ecommerce (15) & brand & \textbf{0.59} & 0.58 & 0.20 & \textbf{0.49} \\
    fortune1000\_2023 (31) & Dropped\_in\_Rank & 0.93 & \textbf{0.94} & 0.68 & \textbf{0.91} \\
    fortune1000\_2023 (31) & Gained\_in\_Rank & 0.73 & \textbf{0.93} & 0.77 & \textbf{0.93} \\
    fortune1000\_2023 (31) & Sector & 0.42 & \textbf{0.89} & 0.25 & \textbf{0.87} \\
    fortune1000\_2023 (31) & HeadquartersState & 0.68 & \textbf{0.88} & 0.65 & \textbf{0.97} \\
    fortune1000\_2023 (31) & Industry  & 0.13 & \textbf{0.23} & 0.12 & \textbf{0.34} \\
    IPM Matches (16) & city & 0.66 & \textbf{0.85} & 0.86 & \textbf{0.94} \\
    shopping trends (19) & Season & 0.22 & \textbf{0.28} & 0.20 & \textbf{0.26} \\
    shopping trends (19) & Category & 0.62 & \textbf{1.00} & 0.68 & \textbf{0.99} \\
    starbucks in california (24) & state & 1.00 & 1.00 & 0.83 & \textbf{1.00} \\
    starbucks in california (24) & city  & 0.04 & \textbf{0.44} & 0.73 & \textbf{0.86} \\
    starbucks in california (24) & county  & 1.00 & 1.00 & 0.88 & \textbf{0.99} \\
    starbucks in california (24) & 24\_hour\_service & 0.00 & \textbf{0.76} & 0.00 & \textbf{0.79} \\
    Restaurants (6) & City & 0.75 & \textbf{0.82} & 0.92 & \textbf{0.97} \\
    BigBasket (14) & category & 0.32 & \textbf{0.92} & 0.91 & \textbf{0.92} \\
    Global PowerPlantDB (40) & source & 0.61 & \textbf{0.90} & 0.84 & 0.84 \\
    AMTRAK (86) & city & 0.57 & \textbf{0.98} & 0.61 &  \textbf{0.81} \\
    Speed Dating (124) & race & 0.30 & \textbf{0.61} & 0.35 & \textbf{0.64}\\
    \bottomrule
    \end{tabular}
    \caption{Data Imputation Task}
    \label{tab:DI_all_datasets}
\end{table*}

\begin{table*}[ht]
    \centering
    \small
    \begin{tabular}{cccccc}
    \toprule
    \textbf{Dataset} & \textbf{Target Column} & \textbf{Baseline} & \textbf{Ours} & \textbf{Baseline} & \textbf{Ours} \\
     (\#columns) & & flan-t5-xxl (11B) & flan-t5-xxl (11B) & Mixtral (8x7B) & Mixtral (8x7B) \\
    \hline
    customer support tickets (17) & Ticket Priority & 0.45 & \textbf{0.93} & 0.86 & \textbf{0.99} \\
    customer support tickets (17) &  Ticket Type & 0.38 & \textbf{0.81} & 0.68 & \textbf{0.98} \\
    customer support tickets (17) &  Ticket Subject & 0.42 &  \textbf{0.68}  & 0.74 & \textbf{0.98} \\
    shopping trends (19) & Category & 0.38 & \textbf{0.76} & 0.90 & \textbf{0.98} \\
    shopping trends (19) & Season & 0.45 & \textbf{0.95} & 0.91 & \textbf{0.96} \\
    GlobalPowerPlantDB (36) & source & 0.34 & \textbf{0.83} & 0.90 & \textbf{0.95} \\
    GlobalPowerPlantDB (36) & country & 0.37 & \textbf{0.85} & 0.83 & \textbf{0.97} \\
    GlobalPowerPlantDB (36) & country long & 0.38 & \textbf{0.87} & 1.00 & \textbf{0.98} \\
    GlobalPowerPlantDB (36) & source & 0.33 & \textbf{0.83} & 0.71 & \textbf{0.95}\\
    flipkart ecommerce (15) & product\_category\_tree & 0.00 & \textbf{0.79} & 0.63 & \textbf{0.70} \\
    flipkart ecommerce (15) & brand & 0.39 & \textbf{0.84} & 0.73 & \textbf{0.87} \\
    finance sentiment analysis (2) & Sentiment & 0.37 & 0.37 & 0.74 & \textbf{0.97} \\
    fortune1000\_2023 (31) & Sector & 0.38 & \textbf{0.39} & 0.98 & \textbf{0.99} \\
    fortune1000\_2023 (31) & HeadquartersState & 0.39 & \textbf{0.89} & 0.99 & 0.99 \\
    fortune1000\_2023 (31) & Industry & 0.38 & \textbf{0.77} & 0.89 & \textbf{0.96} \\
    fortune1000\_2023 (31) & Dropped\_in\_Rank & 0.46 & \textbf{0.86} & 0.90 & \textbf{1.00} \\
    fortune1000\_2023 (31) & Gained\_in\_Rank & 0.42 & \textbf{0.81} & 0.87 & \textbf{0.99} \\
    BigBasket Products (14) & sub\_category & 0.38 & \textbf{0.48} & \textbf{0.80} & 0.45 \\
    BigBasket Products (14) & type & 0.38 & \textbf{0.86} & 0.82 & \textbf{0.88} \\
    BigBasket Products (14) & category & 0.39 & \textbf{0.87} & 0.37 & \textbf{0.95}\\
    Airline Dataset (15) & Continents & 0.43 & \textbf{0.91} & \textbf{0.99} & 0.91 \\
    Airline Dataset (15) & Airport Continent & 0.42 & \textbf{0.76} & 0.91 & \textbf{0.99} \\
    Airline Dataset (15) & Airport Country Code & 0.77 & \textbf{0.91} & 0.98 & \textbf{0.99} \\
    Airline Dataset (15) & Country Name & 0.53 & \textbf{0.96} & 0.95 & \textbf{0.96} \\
    starbucks in california (24) & county & 0.42 & \textbf{0.53} & 1.00 & \textbf{0.98} \\
    starbucks in california (24) & city & 0.40 & \textbf{0.79} & \textbf{0.97} & 0.95 \\
    starbucks in california (24) & 24\_hour\_service & 0.28 & \textbf{0.93} & 0.93 & \textbf{0.99} \\
    starbucks in california (24) & state & 0.38 & \textbf{0.89} & 1.00 & 1.00 \\
    Speed Dating (124) & race & 0.33 & \textbf{0.69} & 0.33 & \textbf{0.73}\\
    IPM Matches (16) & city & 0.42 & \textbf{0.88} & \textbf{0.93} & 0.91 \\
    Adult (15) & (Multiple targets) & 0.50 & \textbf{0.95} & 0.54 & \textbf{0.78} \\
    Hospital (23) & (Multiple targets) & 0.49 & \textbf{0.85} & 0.36 & \textbf{0.81}\\
    \bottomrule
    \end{tabular}
    \caption{Error detection Task}
    \label{tab:ED_all_datasets}
\end{table*}

\begin{table*}[tb!]
    \centering
    \small
    \begin{tabular}{ccccc}
    \toprule                                  
    \textbf{Dataset} & \textbf{Baseline} & \textbf{Ours} & \textbf{Baseline} & \textbf{Ours} \\
    (\#columns) & flan-t5-xxl (11B) & flan-t5-xxl (11B) & Mixtral (8x7B) & Mixtral (8x7B) \\
    \midrule
    Fodors-Zagats (14) & 0.86 &  \textbf{1.00} & 0.92 & \textbf{0.99} \\
    DBLP-GoogleScholar (10) & 0.69 & \textbf{0.85}  &  0.83 & \textbf{0.85} \\
    beers (10) & 0.71 & \textbf{0.89} & 0.88 & \textbf{0.91} \\
    Walmart-Amazon (12) & 0.54 & \textbf{0.87} & 0.91 & 0.91 \\
    iTunes-Amazon (16) & 0.78 & \textbf{0.89} & 0.76 & \textbf{0.91} \\
    DBLP-ACM (8) & 0.90 & \textbf{0.98} & 0.87 & \textbf{0.94}\\
    Amazon-Google (6) & 0.50 & \textbf{0.63} & 0.73 & 0.73 \\
    \bottomrule
    \end{tabular}
    \caption{Entity Matching Task}
    \label{tab:EM_all_datasets}
\end{table*}

\begin{table*}[ht!]
    \centering
    \small
    \begin{tabular}{|p{1cm}|p{3cm}|p{6cm}|p{6cm}|}
        \toprule
        Task & Dataset Name & Manually Selected columns & RL selected columns \\
        \hline
        \multirow{4}{*}{DI}  & Restaurant & Name, Address, Phone & Address, Name \\
        \cline{2-4}
        & BigBasket & sub\_category, product, description, type & description, 
        sub\_category, type, product \\
        \cline{2-4}
        & Global PowerPlant & owner, geo\_source, name, country\_long & geolocation\_source, country, gppd\_idnr \\
        \cline{2-4}
        & AMTRAK & StationName, address1, address2, State, Zip & Zip, StationName, StationServicesPaging \\
        \hline
        \multirow{2}{*}{ED} & Global Power Plant & owner, geo\_source, name, country\_long & geolocation\_source, country, estimated\_generation\_note\_2015 \\
        \cline{2-4}
         & BigBasket & sub\_category, product, description, type  & brand, sub\_category, market\_price \\
        \hline
        \multirow{4}{*}{EM} & Fodors-Zagats & l\_name, l\_addr, l\_city, l\_class, r\_name, r\_addr, r\_city, r\_class & r\_class, l\_class \\
        \cline{2-4}
        & DBLP-GoogleScholar & l\_title, l\_authors, l\_venue, r\_title, r\_authors, r\_venue & r\_title, l\_title, r\_venue, l\_venue \\
        \cline{2-4}
        & beers & l\_Beer\_Name, l\_Brew\_Factory\_Name, l\_Style, r\_Beer\_Name, r\_Brew\_Factory\_Name & l\_Beer\_Name, r\_Beer\_Name, l\_Brew\_Factory\_Name, r\_Brew\_Factory\_Name \\
        \cline{2-4}
        & Walmart-Amazon & l\_title, l\_brand, l\_modelno, r\_title, r\_brand, r\_modelno & r\_title, l\_title, l\_modelno, r\_modelno \\
    \bottomrule
    \end{tabular}
    \caption{Manual and RL selected columns per dataset (the columns that were manually selected and those selected and sequenced by the RL algorithm for 12 datasets across three tasks.)}
    \label{tab:sel_cols}
\end{table*}

\end{document}